\documentclass{article}

\usepackage[final, nonatbib]{neurips_2023}
\usepackage[numbers]{natbib}

\usepackage[utf8]{inputenc} % allow utf-8 input
\usepackage[T1]{fontenc}    % use 8-bit T1 fonts
\usepackage[hidelinks]{hyperref}       % hyperlinks
\usepackage{url}            % simple URL typesetting
\usepackage{booktabs}       % professional-quality tables
\usepackage{amsfonts,amssymb}       % blackboard math symbols
\usepackage{nicefrac}       % compact symbols for 1/2, etc.
\usepackage{microtype}      % microtypography
\usepackage{xcolor}         % colors
\usepackage{colortbl}
\usepackage{bigstrut}
\usepackage{pifont}% http://ctan.org/pkg/pifont
\usepackage{graphicx}
\usepackage{amsmath}
\usepackage{amssymb}
\usepackage{amsthm}
\usepackage{caption}
\usepackage{subfig}
\usepackage{multirow}
\usepackage{float}
\usepackage{wrapfig}

\newcommand{\cmark}{\ding{51}}%

\newcommand{\eg}{\textit{e.g., }}
\newcommand{\ie}{\textit{i.e., }}
\newcommand{\amber}{\textsc{Amber}}
\newcommand{\amberchat}{\textsc{AmberChat}}
\newcommand{\ambersafe}{\textsc{AmberSafe}}
\newcommand{\crystal}{\textsc{CrystalCoder}}
\newcommand{\llm}{\textsc{LLM360}}
\newcommand{\analysis}{\textsc{Analysis360}}

\definecolor{Gray}{gray}{0.95}
\definecolor{DarkGray}{gray}{0.5}
\definecolor{LightCyan}{rgb}{0.88,1,1}
\definecolor{bisque}{rgb}{1.0, 0.89, 0.77}
\definecolor{blanchedalmond}{rgb}{1.0, 0.92, 0.8}
\definecolor{cosmiclatte}{rgb}{1.0, 0.97, 0.91}
\definecolor{cornsilk}{rgb}{1.0, 0.97, 0.86}

\NewDocumentCommand{\hector}{mO{}}{\textcolor{purple}{\textsuperscript{\textit{Hector}}\textsf{\textbf{\small[#1]}}}}

\title{\llm{}: Towards Fully Transparent \\Open-Source LLMs}
\author{%
  Zhengzhong Liu\\
  Petuum \& MBZUAI
  \And
  Aurick Qiao\\
  Petuum
  \And
  Willie Neiswanger\\
  USC \& Petuum
  \And
  Hongyi Wang\\
  CMU
  \And
  Bowen Tan\\
  CMU
  \And
  Tianhua Tao\\
  UIUC
  \And
  Junbo Li\\
  MBZUAI
  \And
  Yuqi Wang\\
  Petuum
  \And
  Suqi Sun\\
  Petuum
  \And
  Omkar Pangarkar\\
  Petuum
  \And
  Richard Fan\\
  Petuum
  \And
  Yi Gu\\
  UCSD
  \And
  Victor Miller\\
  Petuum
  \And
  Yonghao Zhuang\\
  CMU
  \And
  Guowei He\\
  MBZUAI
  \And
  Haonan Li\\
  MBZUAI
  \And
  Fajri Koto\\
  MBZUAI
  \And
  Liping Tang\\
  MBZUAI
  \And
  Nikhil Ranjan\\
  MBZUAI
  \And
  Zhiqiang Shen\\
  MBZUAI
  \And
  Xuguang Ren\\
  MBZUAI
  \And
  Roberto Iriondo\\
  MBZUAI
  \And
  Cun Mu\\
  MBZUAI
  \And
  Zhiting Hu\\
  UCSD
  \And
  Mark Schulze\\
  Petuum
  \And
  Preslav Nakov\\
  MBZUAI 
  \And
  Tim Baldwin\\
  MBZUAI
  \And
  Eric P. Xing\\
  MBZUAI
}

\begin{document}

\maketitle

\begin{figure}[h]
\vspace{-5mm}
\centering
\includegraphics[width=3cm]{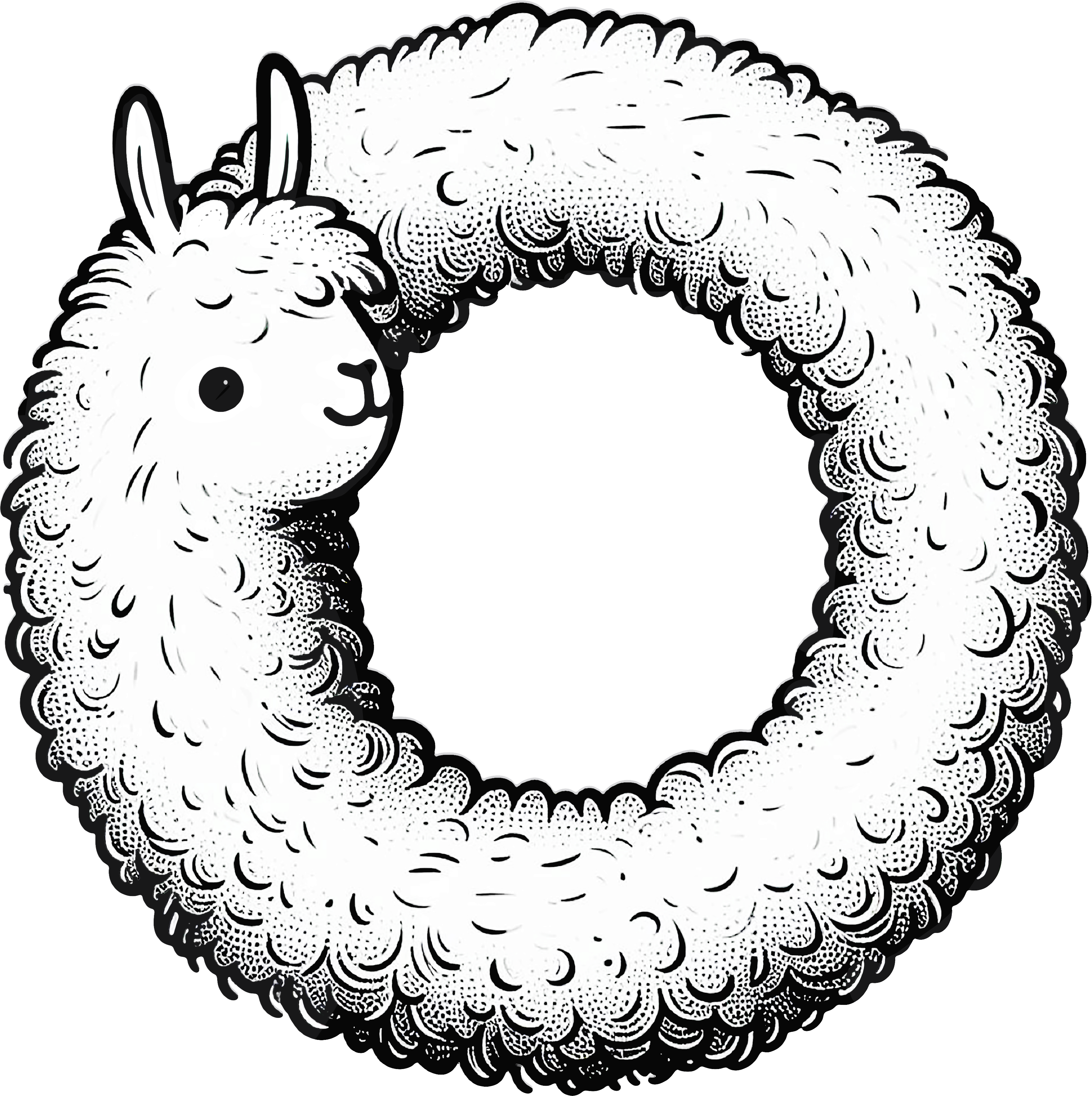}
\vspace{1mm}
\end{figure}

\begin{abstract}
The recent surge in open-source Large Language Models (LLMs), such as LLaMA, Falcon, and Mistral, provides diverse options for AI practitioners and researchers. However, most LLMs have only released partial artifacts, such as the final model weights or inference code, and technical reports increasingly limit their scope to high-level design choices and surface statistics. These choices hinder progress in the field by degrading transparency into the training of LLMs and forcing teams to rediscover many details in the training process. We present \textbf{\llm{}}, an initiative to fully open-source LLMs, which advocates for all training code and data, model checkpoints, and intermediate results to be made available to the community. The goal of \llm{} is to support open and collaborative AI research by making the end-to-end LLM training process transparent and reproducible by everyone. As a first step of \llm{}, we release two 7B parameter LLMs pre-trained from scratch, \amber{} and \crystal{}, including their training code, data, intermediate checkpoints, and analyses (at \href{https://www.llm360.ai/}{\texttt{llm360.ai}}). We are committed to continually pushing the boundaries of LLMs through this open-source effort. More large-scale and stronger models are underway and will be released in the future.
\end{abstract}

\section{Introduction}

%\williex{Para 1: Current LLM landscape; LLMs gaining in popularity and power, closed-source LLMs, recent surge of open-source LLMs}
The landscape of Large Language Models (LLMs) has experienced a remarkable transformation in the past one year, witnessing an unprecedented surge in both the popularity and capabilities of these models. At the forefront of this evolution are proprietary LLMs such as GPT-4~\cite{openai2023gpt4} and Claude~\cite{claude21modelcard}, which have captured the attention of the AI community due to their power and versatility. At the same time, the recent emergence of openly accessible yet highly capable LLMs such as LLaMA~\cite{touvron2023llama,touvron2023llama2}, Falcon~\cite{penedo2023refinedweb}, and Mistral~\cite{jiang2023mistral} allow researchers and practitioners at large to easily obtain, customize, and deploy LLMs in more diverse environments and for more diverse use cases.

%The year 2023 has witnessed several open-source LLM efforts, such as LLaMA, Llama2, MPT, Falcon, JAIS, and more~\cite{touvron2023llama,touvron2023llama2,MosaicML2023Introducing,penedo2023refinedweb,sengupta2023jais}. These works primarily focus on releasing the final model checkpoints of the base models and their instruction-fine-tuned model weights, enabling the development of applications like chatbots and code LLMs based on those released checkpoints\cite{roziere2023code,zheng2023judging}. 

%\williex{Para 2: Issues with current state: Open-Source LLMs are still not transparent enough, and it causes issues that hinder progress; if efforts are \textit{fully transparent} about details of LLM pre-training, fine-tuning, evaluation, and more, then we can truly democratize the study of LLMs, and and reap the benefits of community-wide access to a full-set of information and artifacts.}

Despite the growing influence and accessibility of open-source LLMs, a notable trend has been to restrict visibility and access to their training, fine-tuning, and evaluation processes, including crucial components such as their training code and data. This practice limits the ability of the broader AI research community to study, replicate, and innovate upon advanced LLMs. A more transparent approach to sharing not just the final model but also training details and artifacts is crucial for fostering a more inclusive and collaborative research environment.

Motivated by the above, we note the following specific challenges in LLM research today.

\noindent\textbf{Data Provenance.} Understanding the origins and characteristics of the training data is crucial for assessing the reliability and biases inherent in LLMs. A lack of transparency about data sources and composition hinders the ability to identify and mitigate biases which can be perpetuated in model outputs. Simultaneously, data leakage---where training datasets overlap with benchmark datasets---can lead to misleading performance metrics that obscure a model's general effectiveness (studied in \cite{wei2023skywork,zhou2023dont}). These issues highlight the need for clear documentation of data origins and usage in LLM development.

\noindent\textbf{Reproducibility.} Even with full disclosure of data sources, the lack of access to complete training code, configuration details, and specific datasets can make it challenging to reproduce the results reported in studies. For example, although the training data mixtures are disclosed by LLaMA~\cite{touvron2023llama}, the data processing and training code are not released. Yet, LLMs known to be trained using an open reproduction of LLaMA's data (\eg RedPajama~\cite{together2023redpajama,together2023incite}) still do not fully reproduce its benchmark evaluations~\cite{openlm2023openllama}, indicating that additional data processing or training procedures may be necessary.

\noindent\textbf{Open Collaboration.} The practice of only releasing final model weights not only leads to redundant efforts but also poses uniques challenges in conducting certain research. For instance, research into the emergent abilities of LLMs~\cite{biderman2023emergent,wei2022emergent} or the investigation of how different training data affects model behavior~\cite{yu2023large,xie2023doremi} becomes more challenging without access to intermediate training checkpoints. Researchers are often forced to either work with the final model, which offers limited insights into its developmental nuances, or start from scratch, leading to unnecessary duplication of work and expenditure of compute.

%This practice limits the ability of the broader AI research community to fully understand, replicate, and innovate upon advanced LLMs.
%However, most of the LLM develop details are not fully transparent, only allowing researchers and developers to interact with black-boxes. For instance, LLaMA's technical reports~\hector{this section focuses only on pretraining, need to extend the scope} do not clearly explain the precise methods for constructing pre-training datasets \cite{touvron2023llama,touvron2023llama2}. The open-source community has to make efforts, such as RedPajama~\cite{together2023redpajama}, to reconstruct an equivalent version of LLaMA's pretraining dataset. However, models\hector{citation} trained on these open-source datasets usually fall behind LLaMA in performance, indicating that additional data quality filtering, data cleaning, and data mixing ratio tuning are required to train LLMs with LLaMA-level performance. This lack of transparency in previous open-source LLM efforts can significantly hinder the progress of developers and researchers eager to get involved and contribute.
%\williex{Describe issue with data leakage; unclear if benchmark data is in training sets; reproducibility is limited; researchers cannot access precise mapping between subsets of data and a checkpoint of model progress (e.g. to study training dynamics); especially true for most of the state-of-the-art open sourrce models.}

%\williex{Para 3: Our goal/contributions with LLM360. Summarize the main contributions (e.g. artifacts that we release) and reasons for them.}
\llm{}\footnote{The name \llm{} signifies open-sourcing LLMs from all angles, and that 360 data points (\ie{}checkpoints, data chunks, evaluation results) are released for many of our models.} aims to address the issues above through a comprehensive open-source LLM effort. Models in \llm{} are published with all training and model details (\eg hyperparameters, schedules, architecture, and designs), all intermediate model checkpoints saved during training, and full disclosure of the exact pre-training data used.

Our contributions are:
\begin{itemize}
\item We outline the \llm{} framework, focusing on its design principles and the rationale for fully open-sourcing LLMs. We detail the components of the framework, including datasets, code and configurations, model checkpoints, and training metrics. This framework provides a target for transparency that all present and future \llm{} models strive to meet.
\item We pretrain two new LLMs from scratch and release them under the \llm{} framework. \amber{} is a 7B English LLM pretrained on 1.3T tokens. \crystal{} is a 7B English and code LLM pretrained on 1.4T tokens. We discuss the development details, preliminary evaluations, observations, and lessons we learned from \amber{} and \crystal{}.
\item We release all training code, pretraining data, model checkpoints, and evaluation metrics collected during pretraining for both \amber{} and \crystal{}. Notably, \amber{} is released with 360 model checkpoints saved during training, and \crystal{} with 143.
\end{itemize}

%\williex{Para 4: Full details/description of the contributions and magnitude of what we release; future goals of projects; describe layout of the paper; end with the comparison table}

We aim to make a continuous commitment to fully open-source LLMs by releasing multiple LLMs at various scales. As the first step, in this technical report, we discuss \amber{} and \crystal{}, the first open-source LLMs in the \llm{} series. In the future, we plan to release more pre-trained LLMs that are larger in scale, exhibit better performance, and focus on various domains.

The rest of this report is organized as follows. In \S\ref{sec:related-works}, we discuss related works and the predecessors that inspired \llm{}. In \S\ref{sec:llm360-framework}, we provide a description of the \llm{} framework and the release artifacts that fall into its purview. In \S\ref{sec:model-release}, we discuss the first two LLMs released under \llm{}, \amber{} (\S\ref{sec:amber}) and \crystal{} (\S\ref{sec:crystal}), and preliminary analyses of both. \S\ref{sec:conclusion} concludes.

\section{Related Work}
\label{sec:related-works}

%\williex{Para 5: Comparison with existing efforts, e.g. Pythia}
The closest project to \llm{} is Pythia,
%\hector{the main difference here is that we are advocating to change the way of oss models, instead of designing careful comparison like Pythia, hence our purposes are different}
which also aims at full reproducibility of LLMs~\cite{biderman2023pythia}. The Pythia project provided 154 checkpoints for model sizes from 70M to 12B to better support research on the scaling behavior and learning dynamics of LLMs. While Pythia is a pioneering work, it no longer reflects many recent LLM practices, such as training over trillion-token datasets or training on language and code in different stages. On the other hand, \llm{} defines a release framework prioritizing transparency and reproducibility under which up-to-date models can continue to be released, and our 7B \amber{} model surpasses the 12B Pythia model in public benchmarks~\cite{open-llm-leaderboard}. Overall, Pythia set an early precedent for transparency and reproducibility of LLMs that we aim to perpetuate and expand in \llm{} to modern LLM pretraining regimes.

\begin{table}[ht]
\centering
{\tiny
\begin{tabular}{cccccccccc}
\toprule
\rowcolor{Gray} \textbf{\scriptsize LLM} & \textbf{\scriptsize Release} & \multicolumn{2}{c}{\textbf{\scriptsize Pretraining}} & \multicolumn{2}{c}{\textbf{\scriptsize Checkpoints}} & \multicolumn{3}{c}{\textbf{\scriptsize Pretraining Dataset}} & \textbf{\scriptsize Tokens} \\
\rowcolor{Gray} \textbf{\scriptsize Name} & \textbf{\scriptsize Date} & Code & Config & Model & Optim & Data Mix & Ordering & Available & ($T$)
\bigstrut\\
\midrule
GPT-J~\cite{gpt-j} & May'21 & \cmark & \cmark & \cmark & \cmark & \cmark & \cmark & \cmark & 0.40 \bigstrut\\
GPT-NeoX~\cite{black2022gptneox20b} & Apr'22 & \cmark & \cmark & \cmark & \cmark & \cmark & \cmark & \cmark & 0.40 \bigstrut\\
OPT~\cite{zhang2022opt} & May'22 & \cmark & \cmark & \cmark & & \cmark & & & 0.18 \bigstrut\\
BLOOM~\cite{workshop2022bloom} & Nov'22 & \cmark & \cmark & \cmark & \cmark & \cmark & \cmark & \cmark & 0.34 \bigstrut\\
Pythia~\cite{biderman2023pythia} & Feb'23 & \cmark & \cmark & \cmark & \cmark & \cmark & \cmark & \cmark & 0.30 \bigstrut\\
LLaMA~\cite{touvron2023llama} & Feb'23 & & \cmark & & & \cmark & & & 1.0 \bigstrut\\
%CerebrasGPT~\cite{dey2023cerebras} & Mar'23 & & \cmark & & & \cmark & & \cmark & 0.13 \bigstrut\\
OpenLLaMA~\cite{openlm2023openllama} & May'23 & \cmark & \cmark & \cmark & & \cmark & & \cmark & 1.0 \bigstrut\\
INCITE~\cite{together2023incite} & May'23 & \cmark & \cmark & \cmark & & \cmark & & \cmark & 1.0 \bigstrut\\
MPT~\cite{MosaicML2023Introducing} & May'23 & \cmark & \cmark & & & \cmark & & & 1.0 \bigstrut\\
Falcon~\cite{almazrouei2023falcon} & May'23 & & \cmark & & & \cmark & & & 1.5 \bigstrut\\
Llama 2~\cite{touvron2023llama2} & Jul'23 & & \cmark & & & & & & 2.0 \bigstrut\\
Qwen~\cite{bai2023qwen} & Aug'23 & & \cmark & & & & & & 2.4 \bigstrut\\
Mistral~\cite{jiang2023mistral} & Sep'23 & & & & & & & & ? \bigstrut\\
Yi~\cite{01ai2023yi} & Nov'23 & & & & & & & & ? \bigstrut\\
\rowcolor{cornsilk} \amber & Dec'23 & \cmark & \cmark & \cmark & \cmark & \cmark & \cmark & \cmark & 1.3 \bigstrut\\
\rowcolor{cornsilk} \crystal & Dec'23 & \cmark & \cmark & \cmark & \cmark & \cmark & \cmark & \cmark & 1.4 \bigstrut\\
\bottomrule
\end{tabular}
}
\vspace{5pt}
\caption{\footnotesize Summary of notable open-source LLMs. We note a trend of progressively less disclosure of important pretraining details over time: (1) availability of pretraining code, (2) disclosure of training configurations and hyperparameters, (3) intermediate checkpoints of model weights, (4) intermediate checkpoints of optimizer states, (5) disclosure of data mixture and sources, (6) reproducibility of pretraining data sequence, and (7) availability (or reconstruction scripts) of the pretraining data.}
\label{tab:llm-comparison}
\vspace{-2mm}
\end{table}

In general, open-source LLMs span a wide spectrum of transparency and reproducibility when it comes to their release artifacts. Many recent LLMs only release their final model architecture and weights, keeping their data sources and most training details undisclosed~\cite{touvron2023llama2,bai2023qwen,jiang2023mistral,01ai2023yi}. Some are trained on publicly available datasets~\cite{gpt-j,black2022gptneox20b,workshop2022bloom,biderman2023pythia,openlm2023openllama,together2023incite,shen2023slimpajamadc}, whereas others disclosed their data mixtures but do not make training-ready data available to the public~\cite{zhang2022opt,touvron2023llama,MosaicML2023Introducing,almazrouei2023falcon}. Several LLMs of note have been released with substantially more transparent details and artifacts. For example, EleutherAI models such as GPT-J~\cite{gpt-j} and GPT-NeoX~\cite{gpt-neox-library} included training code, datasets, and up to 150 intermediate model checkpoints. The value of the open-source GPT-NeoX training code was demonstrated by its use in subsequent LLM pretraining by others in the community~\cite{together2023incite,MosaicML2023Introducing}. INCITE~\cite{together2023incite}, MPT~\cite{MosaicML2023Introducing}, and OpenLLaMA~\cite{openlm2023openllama} were released with training code and training dataset, with RedPajama also releasing 10 intermediate model checkpoints.

Overall, we observe a trend that more recent and capable LLMs are becoming more closed in their release artifacts. In contrast, the goal of \llm{} is to release modern and high-quality models while maintaining a high degree of release transparency.

\section{The LLM360 Framework}
\label{sec:llm360-framework}

In this section we present LLM360, a framework for releasing LLMs that promotes open-source transparency, reproducibility, data/model provenance, and collaborative research. LLM360 provides guidance and recommendations for release artifacts that are collected during LLM pre-training and subsequently made publicly available to the community.

As part of the launch of LLM360, we also release two new pre-trained LLMs, which we hope will foster immediate interest and collaboration in the open-source research community. First, \amber{}, an English language LLM with 6.7B parameters trained on 1.25 trillion tokens. Second, \crystal{}, an English and code LLM, also with 6.7B parameters, trained on 1.4 trillion tokens. Details on \amber{} and \crystal{} are reported in \S\ref{sec:model-release}.

\paragraph{Training Dataset and Data Processing Code} The pretraining dataset is the main ingredient of an LLM and significantly impacts its capabilities. Thus, it is important for users and adopters to have visibility into pretraining data to assess potential behavior issues and biases. For example, recent concerns about benchmark data leakage into LLM pretraining is much easier to study when pretraining datasets are available for exploration~\cite{zhou2023dont,wei2023skywork}.

Furthermore, visible pretraining data improves the extensibility of LLMs in later fine-tuning and domain adaptation. Recent work suggests that training on repeated data disproportionately degrades final model performance~\cite{hernandez2022scaling}. Given the breadth of data modern pretraining is performed on, visibility into the original pretraining data is essential for avoiding repeated data in downstream fine-tuning or continued pretraining on specialized domains.

\llm{} advocates for the public release of the data LLMs are pretrained on. When applicable, details about data filtering, processing, and training order should be released as well. Doing so equips the community with better tools to assess the capabilities and risks of LLMs and to reproduce and build upon existing LLMs for future use cases.

%\williex{TODO: goals are to release all the data, the order of the data, how order maps to model checkpoints, weights of domains, whether there is repetition; it's important/necessary to be able to check for data leakage. Also important include all data processing code and tokenization. Also is important to disclose filtering, data engineering, train/test split, fill-in-the-middle details.}

\paragraph{Training Code, Hyperparameters, and Configurations}
\begin{wrapfigure}{r}{0.25\textwidth}
    \vspace{-4mm}
    \centering
    \includegraphics[width=0.25\textwidth]{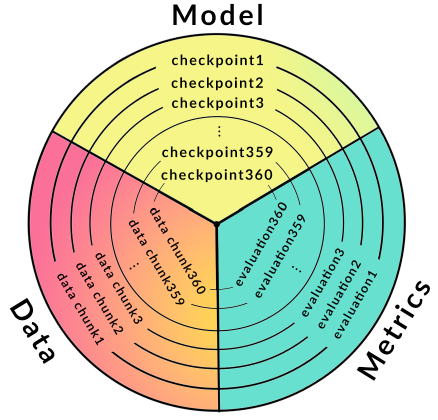}
    \includegraphics[width=0.20\textwidth]{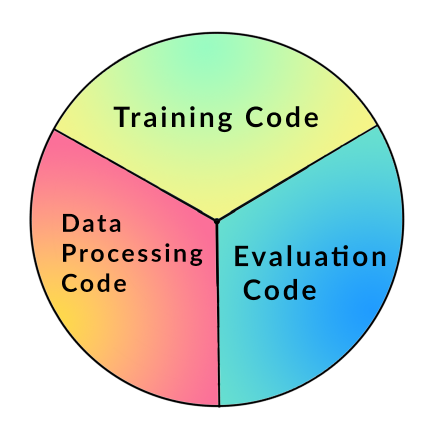}
    \caption{Artifacts relea- sed by the \llm{} project include data chu- nks, model checkpoints, and metrics, at over 360 time stamps of training (and code for all parts).}
    \label{fig:llm360overview}
    \vspace{-12mm}
\end{wrapfigure}
%LLM pre-training is clearly a computationally daunting task, requiring a massive amount of computing resources, \eg GPUs, and extensive optimizations in distributed training systems. Moreover, improper training configurations and hyperparameters can easily cause the training to fail, \eg loss divergence. Existing LLM pre-training frameworks either lack support for all possible parallelism strategies (\eg FSDP solely without full data, tensor-model, and pipeline parallelism) or only support a few types of LLM architectures~\cite{MosaicML2023Introducing,shoeybi2019megatron}.
These code and settings have a significant impact on the performance and quality of LLM training, and are not always publicly disclosed.
For example, we observed that a carefully balanced hybrid data-model-pipeline (3D) parallelism~\cite{narayanan2021efficient} can outperform the standard FSDP in PyTorch by up to 15\% on our Nvidia A100 clusters.
%Moreover, specific training code, configurations, and hyperparameters are usually not fully transparent~\cite{touvron2023llama,touvron2023llama2}.
Another example we observed is that it is essential to keep the inverse frequency matrix in RoPE positional embedding in FP32~\cite{su2021roformer}, which aligns with the observation in Qwen~\cite{bai2023qwen}.

In \llm{}, we open-source all our LLM pre-training frameworks, hyperparameters, as well as the configurations. These include the entire training source code, training parameters such as learning rates and batch sizes, and system configurations such as parallelism dimensions. %We hope that these artifacts will foster reproducible training and help the community understand the training behavior as well as system performance.

\paragraph{Model Checkpoints} It is typical during LLM training to periodically save checkpoints of the model to persistent storage. These checkpoints are not only crucial for recovery from faults during training, but also useful in post-training research such as studying different data and/or hyperparameter schedules, or reproducing infrequently-occurring training faults (\eg loss spikes, NaN results). Recent research on model quantization and compression heavily relies on analysis of model weights and the dynamics during training~\cite{dettmers2022llmint8,liu2023llmqat}.
%\hector{can mention simple analysis like weight distribution llm.int8() can provide insights and have impact over research areas like model compression, quantilization}

\llm{} models are published with all intermediate checkpoints saved during their training, including model weights and optimizer states (when applicable, \eg Adam~\cite{kingma2017adam} moving averages). These checkpoints enable continued training from a range of starting points without training from scratch, making it easier to study and reproduce a wider variety of effects during training.

\paragraph{Metrics} LLMs undergo training over weeks to months, and the trends and evolution patterns over this training period can offer valuable information. However, access to detailed logs and intermediate metrics for LLMs is currently limited to groups involved in pretraining, hindering a comprehensive study of LLMs. These statistics often contain key insights that cannot be directly derived otherwise, and even a simple analysis on the metrics, such as computing metric variances or norms, can reveal significant findings. For instance, the team behind GLM proposed an effective gradient shrinking algorithm for handling loss spikes and NaN losses by analyzing gradient norm behaviors~\cite{zeng2023glm130b}.

Our aim with \llm{} is to alleviate this problem by completely open sourcing the logs and metrics we collect. This includes system statistics (e.g., GPU workload), training logs (e.g., loss, gradient norm), and evaluation metrics (e.g., perplexity, downstream tasks). Access to these logs may facilitate a deeper understanding of the whole training process, including how LLMs evolve during various training scenarios. We provide easy access to the figures by sharing directly on the LLM360 Weights~\&~Biases page\footnote{\url{https://wandb.ai/llm360/projects}}. A few example metrics include downstream evaluation results, training loss, gradient norm, etc.
% Figure~\ref{fig:4_plots} shows a few examples metrics including downstream evaluation results, gradient norm, etc.

% \begin{figure}[ht]
%     \centering
%     \includegraphics[width=0.8\linewidth]{figs/4_plots.pdf}
%     \caption{Example figures shared on LLM360's public Weights and Biases page}\label{fig:4_plots}
% \end{figure}

In \S\ref{sec:analysis360}, we introduce how one can make use of the metrics, and illustrate an experiment tracking the memorization behavior of a model throughout training. The metrics are released in coordination with the data chunks and checkpoints for researchers to easily find their correspondence. Furthermore, we provide open access to the analysis and evaluation code used  to foster reproducibility. The code and all the metrics can be found at an \llm{} repository: \href{https://github.com/llm360/analysis360}{Analysis360}.

%            - Describe why logs and metrics are helpful (and why releasing them over each ckpt is important).

%             - Mention prior work (e.g. Pythia) also showed value in this.
%             - Mention how we can see variance of metrics to know if a value is noise/statistically significant.
%             - Describe teh different analyses that we do.

\section{Initial Model Release}
\label{sec:model-release}

% \williex{Give a sentence/paragraph here that introduces the (moved) Table 4, which will include the full set of Amber and CrystalCoder results. Then give a reference to the new table.}

% \williex{Next we will provide details for our two initial released models.}

\subsection{Amber}
\label{sec:amber}
\begin{wrapfigure}{r}{0.25\textwidth}
    \vspace{-15mm}
    \centering
    \includegraphics[width=0.25\textwidth]{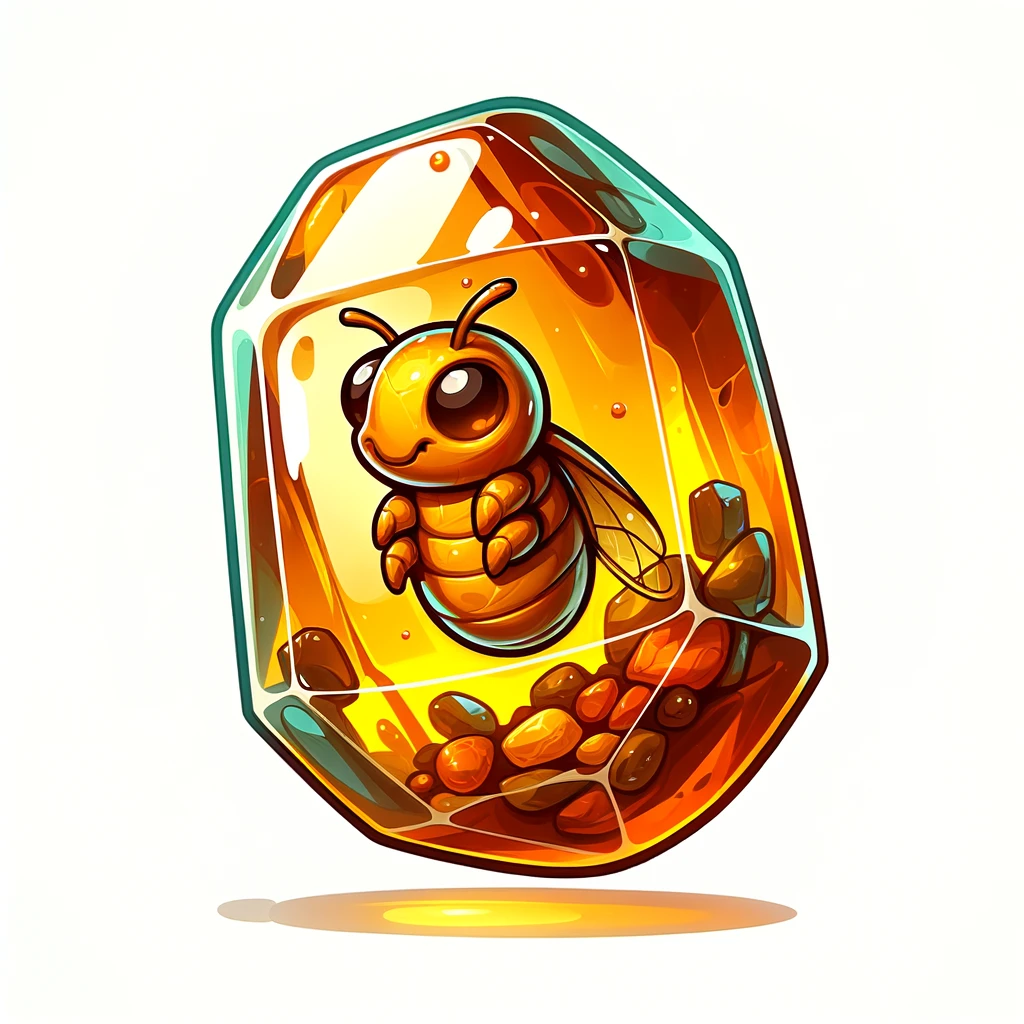}
    \caption{\amber~is a 7B parameter English open-source LLM.}
    \label{fig:amberlogo}
    \vspace{-10mm}
\end{wrapfigure}
In this section, we introduce \amber{}, the first model in the \llm{} family, as well as the finetuned models \amberchat{} and \ambersafe{}.

\subsubsection{Details on Data Preparation and Model Architectures}

Below we review the details of our pre-training dataset, including data preprocessing, format, data mixing ratios, along with architectural details of our LLM model and specific pre-training hyperparameters. The exact setup of \amber{} can be found in the \href{https://github.com/llm360}{LLM360 code base}.

\paragraph{Details on our pre-training dataset} We conduct the data preparation process similar to OpenLLaMA\footnote{\url{https://github.com/openlm-research/open_llama\#dataset-and-training}}. Specifically, our pretraining data is a mixture of RefinedWeb, StarCoder, and RedPajama-v1. A slight difference with OpenLLaMA-v2 is our inclusion of C4, since we do not intend to introduce dupliciated documents after the deduplication process conducted by RefinedWeb. We simply put together all the original aforementioned datasets (without any further cleaning, filtering, or sub-sampling), conduct a global permutation, and partition them evenly into 360 data chunks. In total, we have 1.26 Trillion tokens. Table \ref{tab:data_mix} presents the combination.
%We follow the data preperation process described in OpenLLaMA-v2 \footnote{\url{https://github.com/openlm-research/open_llama\#dataset-and-training}}. Specifically, our pretraining data is a mixture of RefinedWeb, StarCoder, and RedPajama-v1 without CommonCrawl. A slight difference with OpenLLaMA-v2 is our inclusion of C4. We simply put together all original aforementioned datasets (without any further cleaning, filtering, or sub-sampling), conduct a global permutation, and partition them evenly into 360 data chuncks. In total, we have 1.26 Trillions of tokens. Table \ref{tab:data_mix} presents the combination.

\paragraph{The LLM architecture}
We used the exact same model architecture as LLaMA 7B\footnote{The architectural details are directly fetched from \url{https://huggingface.co/huggyllama/llama-7b}}. Detailed LLM architectural configurations are summarized in Table~\ref{tab:7b-config}, incorporating rotary positional embeddings (RoPE) at each layer of the network~\cite{su2021roformer}.

\paragraph{Pre-training procedure and hyperparameters}
We followed the pre-training hyperparameters
from LLaMA as closely as possible~\cite{touvron2023llama}.
\amber{} is trained using the AdamW optimizer with the following hyperparameters: $\beta_1=0.9, \beta_2=0.95$. The initial learning rate is set to $\eta=3e^{-4}$, following a cosine learning rate schedule that decreases to a final rate of $\eta=3e^{-5}$. We apply a weight decay of $0.1$ and use gradient clipping at $1.0$. The model is warmed up over $2,000$ steps. Differing from the LLaMA setup, based on our hardware setting with 224 GPUs, we use a pre-training batch size of $2,240$ ($224 \times 10$) instead of $2,048$.

\begin{minipage}[t]{.48\textwidth}
\begin{table}[H]
    \centering
{\footnotesize
    \begin{tabular}{ll}%{@{}r|c@{}}
\toprule
\rowcolor{Gray} Subset        & Tokens (Billion) \bigstrut\\ \midrule
Arxiv         & 30.00                                \bigstrut\\
Book          & 28.86                                \bigstrut\\
C4            & 197.67                               \bigstrut\\
Refined-Web   & 665.01                               \bigstrut\\
StarCoder     & 291.92                               \bigstrut\\
StackExchange & 21.75                                \bigstrut\\
Wikipedia     & 23.90                                \bigstrut\\ \midrule
Total         & 1259.13          \bigstrut\\ \bottomrule
\end{tabular}}
\vspace{5pt}
\caption{Data mix in \amber{} pre-training.}
    \label{tab:data_mix}
\end{table}
\end{minipage}
\hfill
\begin{minipage}[t]{.48\textwidth}
\begin{table}[H]
\centering
{\footnotesize
\begin{tabular}{ll}
\toprule
\rowcolor{Gray} Hyperparameter & Value  \bigstrut\\
\midrule
Number Parameters & 6.7$B$ \bigstrut\\
Hidden Size & 4096 \bigstrut\\
Intermediate Size (in MLPs) & 11008 \bigstrut\\
Number of Attention Heads & 32 \bigstrut\\
Number of Hidden Layers & 32 \bigstrut\\
RMSNorm $\epsilon$ & $1e^{-6}$ \bigstrut\\
Max Seq Length & 2048 \bigstrut\\
Vocab Size & 32000 \bigstrut\\
\bottomrule
\end{tabular}}\\
\vspace{10pt}
\caption{LLM architecture \& hyperparameters.}
\label{tab:7b-config}
\end{table}
\end{minipage}

\subsubsection{Details on the Pre-training Infrastructure}
\begin{wrapfigure}{r}{0.5\textwidth}
    \vspace{-5mm}
    \centering
    \includegraphics[width=0.5\textwidth]{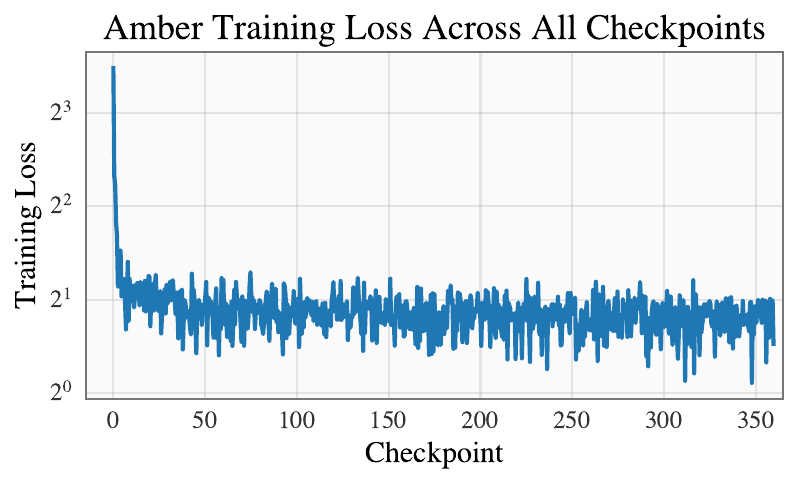}
    \caption{The training loss of \amber{} over all model checkpoints.}
    \label{fig:enter-label}
    \vspace{-10mm}
\end{wrapfigure}

\amber{} is trained on an in-house GPU cluster.

\paragraph{The GPU cluster}
The GPU cluster consists of 56 DGX A100 nodes, each equipped with $4\times$ 80GB A100 GPUs. Each GPU is connected with 4 links NVLink. Cross node connection setting is 2 port 200 Gb/sec (4$\times$ HDR) InfiniBand. The throughput we manage to achieve with our distributed training framework is around 582.4$k$ tokens per second. %It is important to note that our DGX nodes are configured without NVLink connectivity within each node, thus somewhat constraining our pre-training throughput due to limited communication bandwidth.

\paragraph{The pretraining framework} Our pretraining framework is lit-llama\footnote{\url{https://github.com/Lightning-AI/lit-llama}} developed based on PyTorch Lightning. We used mixed-precision during pre-training with BF16 for activations and gradients and FP32 for model weights~\cite{micikevicius2017mixed}.

\subsubsection{Finetuned \amber{} models}
We also release a few finetuned versions of \amber{}, namely \amberchat{} and \ambersafe{}. \amberchat{} is trained on \href{https://huggingface.co/datasets/WizardLM/WizardLM_evol_instruct_V2_196k}{the evolved instruct training data} as used by WizardLM~\cite{xu2023wizardlm}. We use FastChat~\cite{zheng2023judging} to finetune the model for 3 epochs on 8 A100s (80G) distributed by FSDP~\cite{zhao2023pytorch}, the learning rate is $2\times 10^{-5}$, gradient accumulation steps is $16$, warmup ratio is $0.04$.  We also finetune an aligned version of the model: \ambersafe{}, by conducting Direct Parameter Optimization (DPO)~\cite{rafailov2023direct}. \ambersafe{} is trained on \href{https://huggingface.co/datasets/icybee/share_gpt_90k_v1}{ShareGPT 90K}\footnote{The base model for this is checkpoint 355 instead of the last checkpoint}, and further optimized on the SafeRLHF dataset~\cite{ji2023beavertails}. We set $\beta$ to 0.1, gradient accumulation steps to 4, and the learning rate to $5\times 10^{-7}$.
\subsubsection{Results and Analysis}

\begin{figure}[ht]
    \centering
    \includegraphics[width=0.49\textwidth]{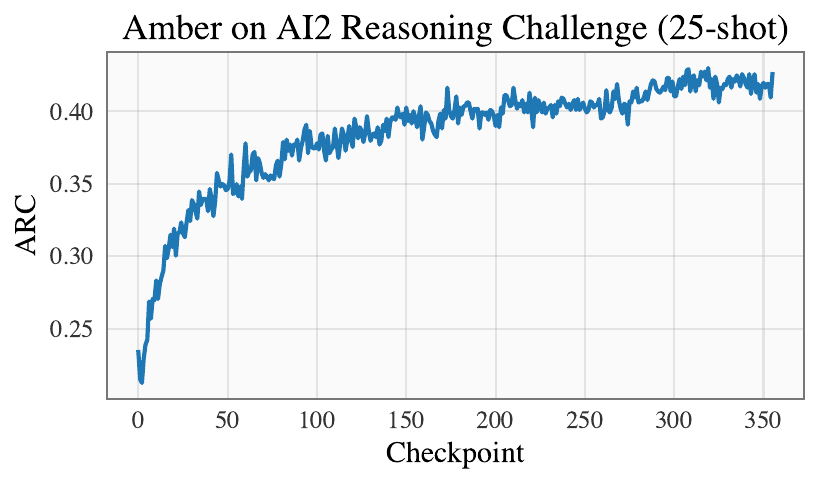}
    \includegraphics[width=0.49\textwidth]{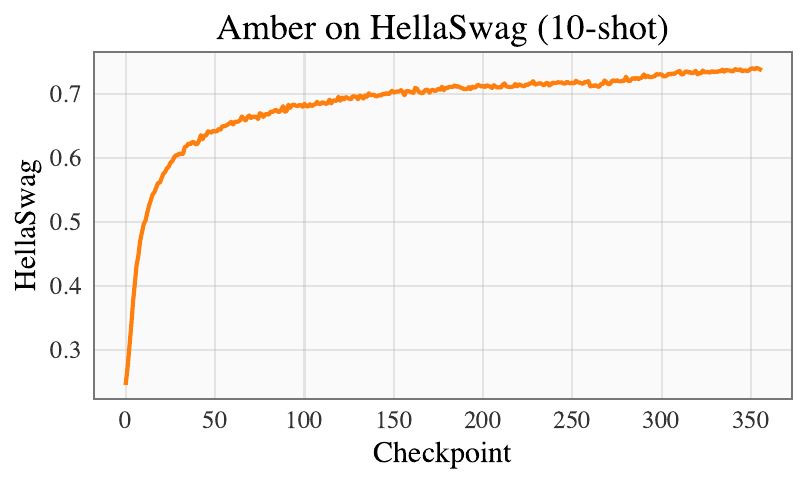}
    \includegraphics[width=0.49\textwidth]{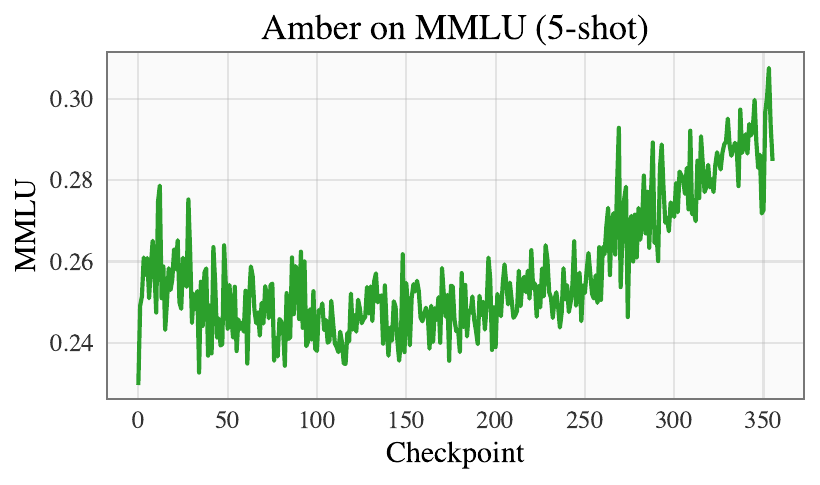}
    \includegraphics[width=0.49\textwidth]{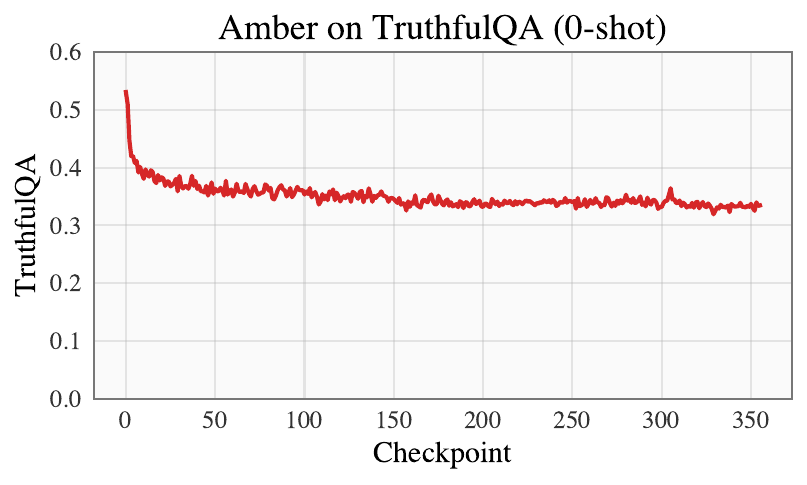}
    \caption{Results for \amber{} on the Open LLM leaderboard metrics.}
    \label{fig:openllm}
\end{figure}

\paragraph{Benchmark Results}
We use four benchmark datasets in the Open LLM Leaderboard\footnote{\url{https://huggingface.co/spaces/HuggingFaceH4/open_llm_leaderboard}} as our evaluation on different aspects, \ie ARC, HellaSwag, MMLU, and TruthfulQA, following the leaderboard settings. We run the evaluation on all 360 checkpoints, to observe the model ability across the pretraining process. As shown in Figure \ref{fig:openllm}, we can see that the HellaSwag and ARC evaluation scores monotonically increase during pre-training, while the TruthfulQA score seems to decrease as the training proceeds. Another interesting trend is observed in the MMLU progress, where the score decreases in the initial stage of pretraining and then starts to increase.

%\paragraph{High precision Ablation.}
%%We conduct an ablation study to see if using higher precision helps with the pretraining. Specifically, we run our last 40 data chunks over 360 of them using 32-bit full precision instead of mixed precision.

In Table~\ref{tab:amber-llm-comparison}, we compare the final model performance of \amber{} to a set of models trained around similar time, namely OpenLLaMA, RedPajama-INCITE, Falcon, MPT. Many are inspired by the design of LLaMA. We found that \amber{} is relatively competitive in scores such as MMLU, but its performance on ARC is behind the curve. We also find that our finetuned \amber{} models are relatively strong, even compared with other similar models. In our early study, we note that \amberchat{} simply trained on ShareGPT 90K also demonstrates much higher performance than our base model, which is slightly different from the trends shown on other models in the table. We leave further investigation of this to future work.  

\begin{table*}[ht]
\centering
{\scriptsize
\begin{tabular}{ccccc|c}
\toprule
\rowcolor{Gray} The LLMs & ARC & HellaSwag & MMLU & TruthfulQA & Avg. \bigstrut\\
\midrule
 LLaMA2-7B-chat & 52.9  & 78.55 & 48.32 & 45.57 & 56.34 \bigstrut\\
 LLaMA2-7B & 53.07  & 77.74 & 43.8 & 38.98 & 53.39 \bigstrut\\
\rowcolor{cornsilk} \textsc{AmberSafe} & 45.22 & 74.14 & 37.78 & 55.44 & 53.15 \bigstrut\\
 LLaMA-7B & 50.94  & 77.8 & 35.67 & 34.34 & 49.69 \bigstrut\\
\rowcolor{cornsilk} \textsc{AmberChat} & 42.83 & 74.03 & 38.88 & 40.72 & 49.12 \bigstrut\\
 OpenLLaMA-v2-7B & 43.69  & 72.2 & 41.29 & 35.54 & 48.18 \bigstrut\\
 MPT & 47.7 & 77.57 & 30.8 & 33.44 & 47.38 \bigstrut\\
 Falcon-7B & 47.87  & 78.13 & 27.79 & 34.26 & 47.01 \bigstrut\\
 RedPajama-INCITE-7B-Instruct & 44.11 & 72.02 & 37.61 & 33.96 & 46.93 \bigstrut\\
 Falcon-7B-instruct & 46.16 & 70.85 & 25.66 & 44.07 &  46.69 \bigstrut\\
 OpenLLaMA-v1-7B & 47.01  & 71.98 & 30.49 & 34.85 & 46.08 \bigstrut\\
\rowcolor{cornsilk} \amber & 41.89  & 74.14 & 30.76 & 34.00 & 45.20 \bigstrut\\
 RedPajama-INCITE-7B-Base & 46.25  & 71.63 & 27.68 & 33.03 & 44.65 \bigstrut\\
 RedPajama-INCITE-7B-Chat & 42.06  & 70.82 & 26.94 & 36.09 & 43.98 \bigstrut\\
% Pythia-6.7B & 40.1  & 65 & 24.64 & 32.85 & 40.65 \bigstrut\\
\bottomrule
\end{tabular}}
\caption{Open LLM leaderboard comparisons for a few LLMs developed around the same time.}
\label{tab:amber-llm-comparison}
\end{table*}

\NewDocumentCommand{\gy}{ mO{} }{\textcolor{purple}{\textsuperscript{\textit{Yi's Note}}\textsf{\textbf{\small[#1]}}}}
\subsubsection{Issues Encountered During Pre-training}
In this section, we discuss several major issues encountered during the pre-training process of \amber. These issues could potentially impact our final model performance. We have addressed most of these issues in subsequent LLM pre-training efforts.

\vspace{-1mm}
\paragraph{NaN loss on a few data chunks} During the pre-training procedure, we encountered NaN loss in four out of 360 data chunks. Whenever we faced this issue, we tentatively skipped the entire data chunk. Initially our plan was to train on these four data chunks in later stage of the training, however, we found that these data chunks tend to cause NaN loss regardless of the position of training. We end up finishing our training by taking the first four chunks from the training sequence to complete our learning rate schedule.

\vspace{-1mm}
\paragraph{Missing optimizer states} In our pre-training framework, we did not manage to save the optimizer states; we only saved model checkpoints for each data chunk. This oversight might be the cause of the NaN loss issue observed in the four data chunks, as mentioned earlier. Each time we resumed pre-training from a previous model checkpoint, the optimizer state in the AdamW optimizer was re-initialized. This re-initialization could potentially affect model training stability.

\vspace{-1mm}
\paragraph{Discrepancies on the precision of checkpoints} In the initial phase of pre-training, our codebase had an issue where model checkpoints were saved with BF16 precision, despite our mixed precision training process maintaining model weights at FP32. This issue was later identified and rectified by our team, ensuring that all subsequent model checkpoints were saved with FP32 precision. We anticipate that the initial BF16 model checkpoints may have contributed to some degree of accuracy drop in the model.

\label{sec:crystal}
\begin{wrapfigure}{r}{0.23\textwidth}
    \vspace{-4mm}
    \centering
    \includegraphics[width=0.23\textwidth]{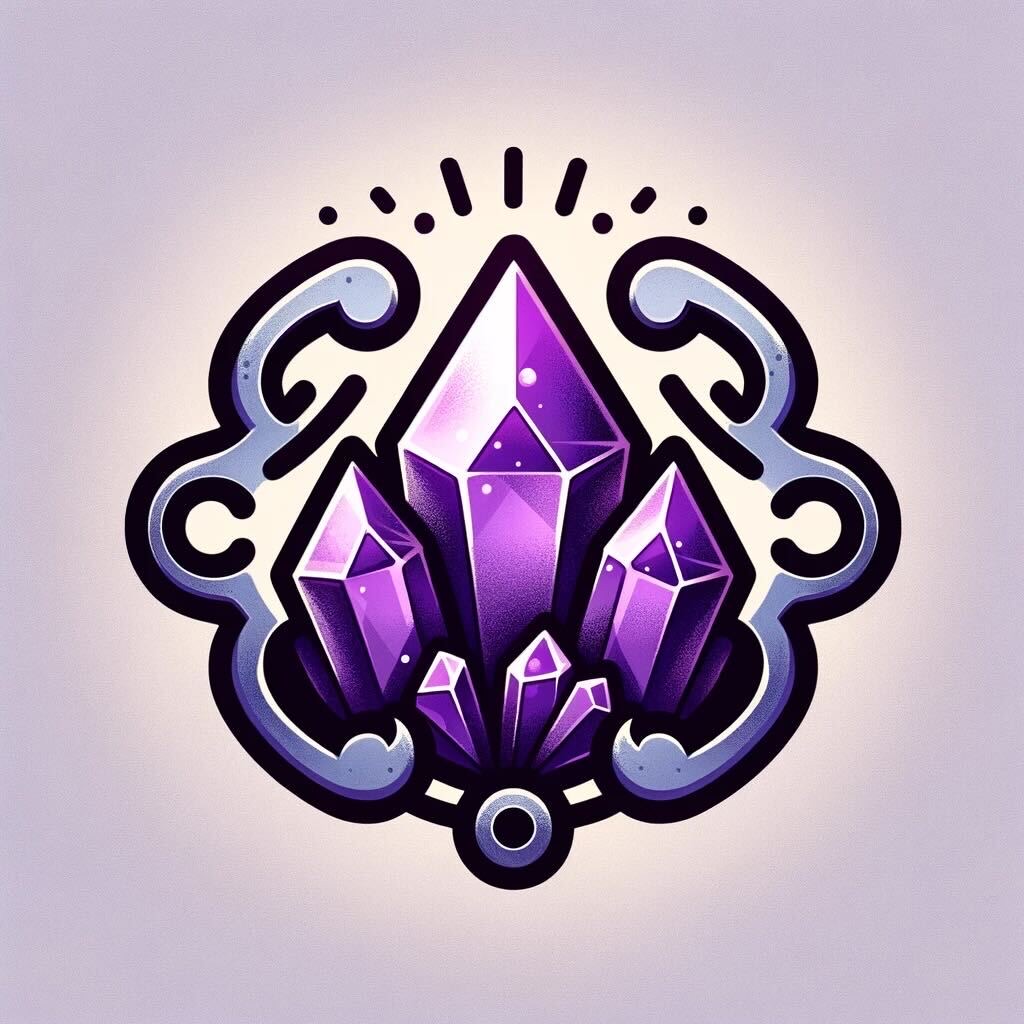}
    \caption{\crystal{} is a 7B parameter English and code open-source LLM.}
    \label{fig:cclogo}
    \vspace{-5mm}
\end{wrapfigure}
\subsection{\crystal{}}
This section provides a summary of the dataset and the model architecture utilized in \crystal{}. For a detailed evaluation of results on benchmarks and a comparison with previous works on specific benchmarks, we refer readers to our future reports.

\vspace{-1mm}
\paragraph{3-Stage Pre-training Dataset}
The pre-training dataset employed in \crystal{} is a blend of SlimPajama \cite{cerebras2023slimpajama} and StarCoder data \cite{li2023starcoder} with around 1382B tokens in total. Diverging from previous approaches such as Code Llama \cite{roziere2023code}, which strictly sequentially trains on English and coding data, we adopt a more gradual approach by seamlessly combining and training on both types of data, to provide a balance between code and general ability. The training process is divided into three stages. In the first stage, we train on half of the SlimPajama data, totaling around 345 billion tokens. Moving to the second stage, the remaining half of the SlimPajama data is utilized, along with two epochs of StarCoder data, resulting in approximately 927 billion tokens. In the third stage, we train on Python and web-related data, encompassing HTML, JavaScript, and CSS subsets from StarCoder, totaling 100 billion tokens. Additionally, we sample 10 billion tokens from the SlimPajama dataset in this stage. The preprocessed data and data mixing scripts are released in the Huggingface and Github repository of \llm{}.

\begin{figure}[ht]
    \centering
    \includegraphics[width=0.49\textwidth]{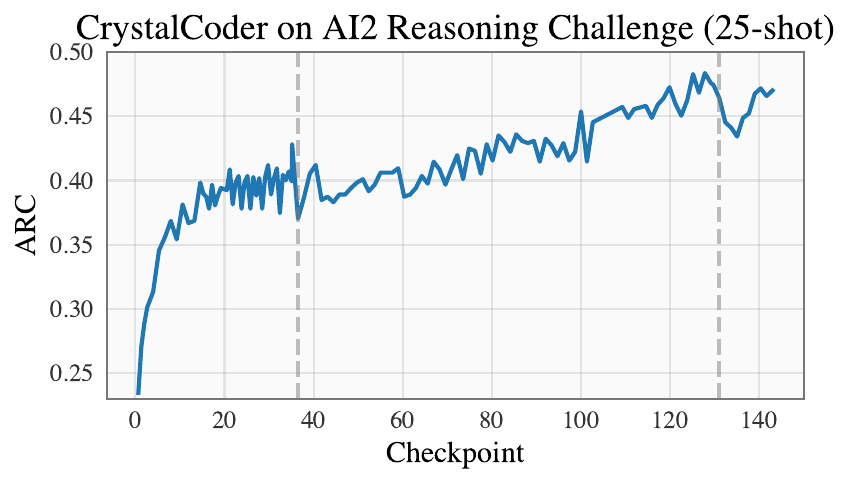}
    \includegraphics[width=0.49\textwidth]{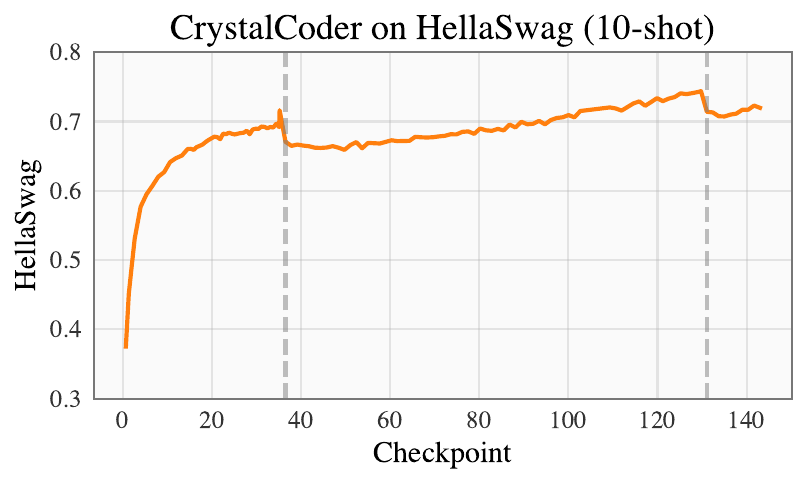}
    \includegraphics[width=0.49\textwidth]{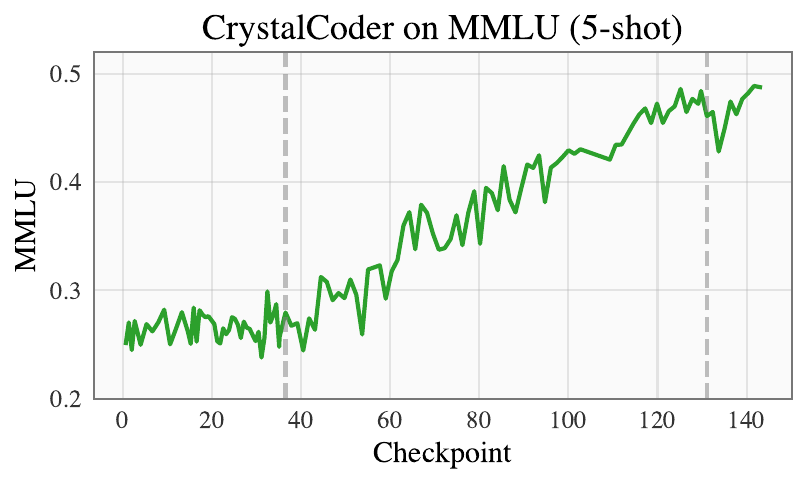}
    \includegraphics[width=0.49\textwidth]{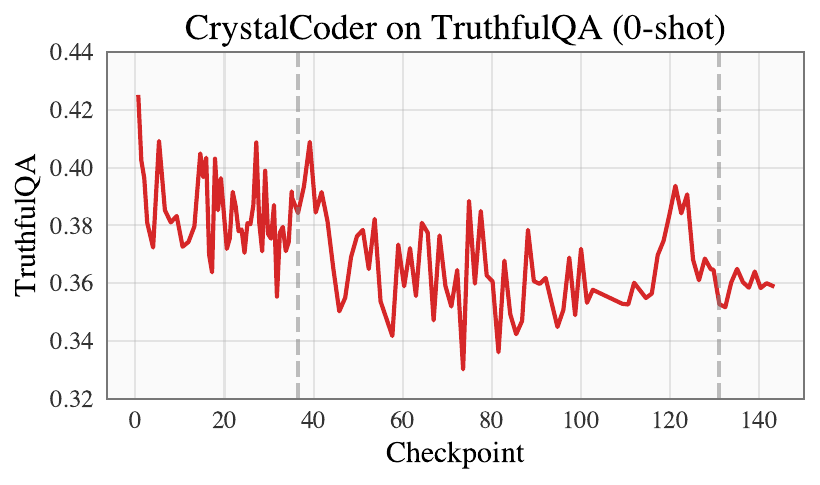}
    \includegraphics[width=0.49\textwidth]{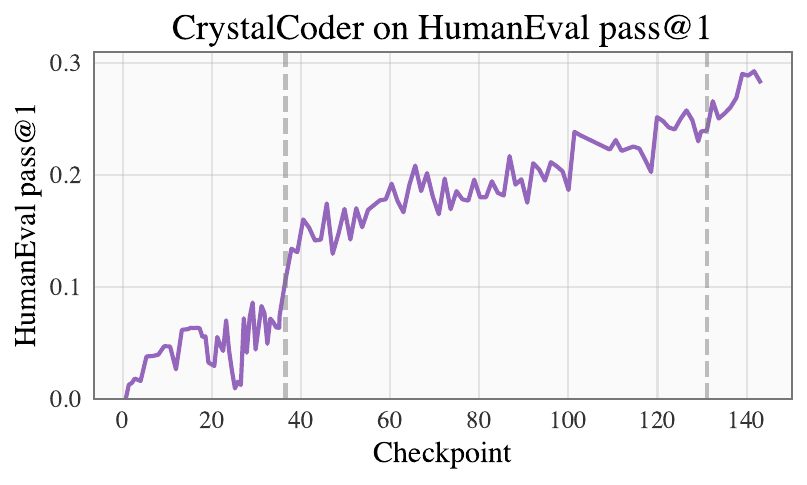}
    \includegraphics[width=0.49\textwidth]{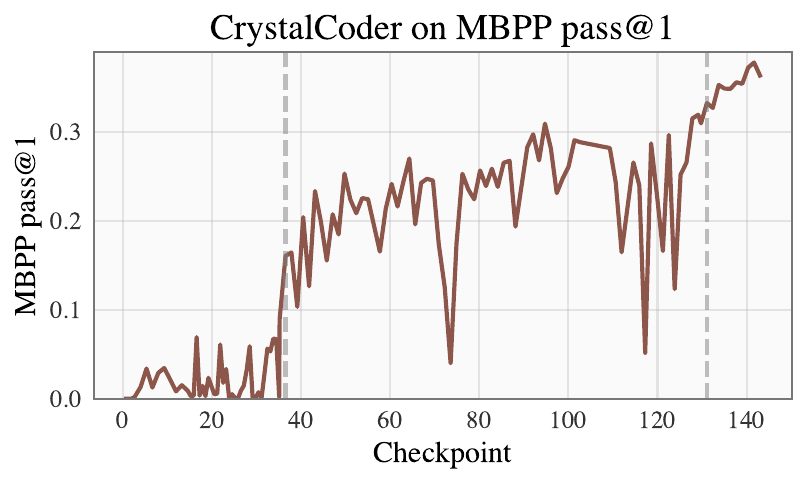}
    \vspace{-2mm}
    \caption{Results for \crystal{} on the Open LLM leaderboard metrics. Grey vertical dashed lines denote the transition between the three stages of training.}
    \label{fig:ccresults}
    \vspace{-5mm}
\end{figure}

\vspace{-1mm}
\paragraph{Model Architecture}
\crystal{} employs a model architecture closely resembling LLaMA 7B, with the incorporation of maximal update parameterization (muP) \cite{yang2022tensor}. In addition to this specific parameterization, we have made several slight modifications, the application of RoPE restricted to the first 25\% of hidden dimensions (similar to the implementation of GPT-NeoX~\cite{gpt-neox-library}), and the use of a sequence length of 2048 with an embedding dimension of 32032. In addition, we simply use LayerNorm instead of RMSNorm since the CG-1 architecture supports efficient computation for vanilla LayerNorm.

\paragraph{Compute Infrastructure}
\crystal{} is trained on the Cerebras Condor Galaxy 1 (CG-1), a 4 exaFLOPS, 54 million core, 64-node cloud AI supercomputer\footnote{\url{https://www.cerebras.net/condor-galaxy-1}}.

\paragraph{Open LLM Leaderboard and Code Evaluations}
We also benchmark this model on the four benchmark datasets in the Open LLM Leaderboard (similar to \amber), as well as coding benchmark datasets, including HumanEval pass@1, and MBPP pass@1.
We show results in Figure~\ref{fig:ccresults}.

\begin{table*}[ht]
\centering
{\scriptsize
\begin{tabular}{c|ccccc|ccc|c}
\toprule
\rowcolor{Gray} \textbf{The LLMs} & \multicolumn{5}{c}{\textbf{Language Tasks}} & \multicolumn{3}{c}{\textbf{Code Tasks}} & \textbf{Avg.}  \bigstrut\\
\rowcolor{Gray} & ARC & HellaSwag & MMLU & TruthfulQA & Avg. & HumanEval & MBPP & Avg. & \\
\midrule
Mistral-7B & 59.98 & 83.31 & 64.16 & 42.15 & 63.40 & 29.12 & 38.78 & 33.95 & 48.68 \bigstrut\\
\rowcolor{cornsilk} \crystal{} (7B) & 47.01  & 71.97 & 48.78 & 35.91 & 50.92 & 28.38 & 36.38 & 32.38 & 41.65 \bigstrut\\
CodeLlama-7B & 39.93 & 60.80 & 31.12 & 37.82 & 42.42 & 33.50 & 41.40 & 37.45 & 39.94 \bigstrut\\
OpenLLaMA-v2-7B & 43.69  & 72.20 & 41.29 & 35.54 & 48.18 & 15.32 & 12.69 & 28.01 & 38.10 \bigstrut\\
LLaMA2-7B & 53.07 & 77.74 & 43.80 & 38.98 & 53.39 & 13.05 & 20.09 & 16.57 & 34.98 \bigstrut\\
LLaMA-7B & 50.94 & 77.80 & 35.67 & 34.34 & 49.69 & 10.61 & 17.04 & 13.83 & 31.76 \bigstrut\\
Falcon-7B & 47.87  & 78.13 & 27.79 & 34.26 & 47.01 & 9.42 & 13.39 & 11.41 & 29.21 \bigstrut\\
StarCoder-15B & -- & -- & -- & -- & -- & 33.63 & 43.28 & 38.46 & -- \bigstrut\\
\bottomrule
\end{tabular}}
\caption{Evaluation comparisons among a few notable code and language models. The last column is the average of the language task average and the code task average. \crystal{} strikes a good balance between both language and code tasks.}
\label{tab:llm-comparison}
\end{table*}

\begin{figure}
    \centering
    \includegraphics[width=\textwidth]{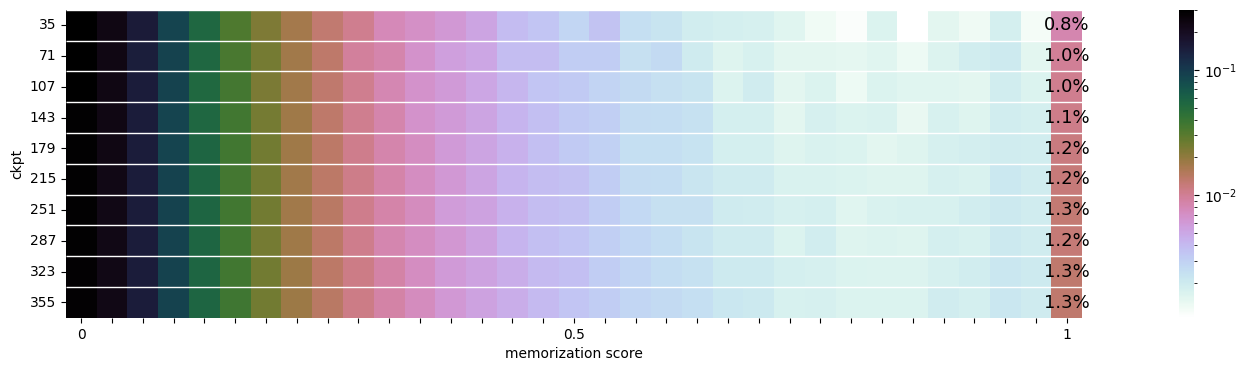}
    \caption{Each row corresponds to the distribution of memorization scores of a checkpoint. We annotate the percentage of $\mathrm{score}=1$ ($k$-extractible) for clearer demonstration.}
    \label{fig:mem-overall}
    \vspace{-8mm}
\end{figure}

\begin{minipage}[b]{.4\textwidth}
\begin{figure}[H]
    \centering
    \includegraphics[height=1.3in]{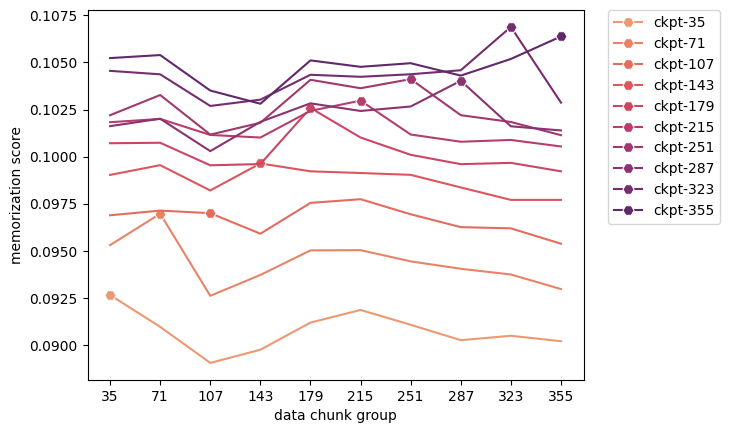}
    \captionof{figure}{Memorization score on data chunk for each checkpoint. The marked spots indicate the latest chunk seen by that checkpoint. The part on right of each mark indicates unseen data.}
    \label{fig:mem-data}
\end{figure}
\end{minipage}%
\hspace{2mm}
\begin{minipage}[b]{.55\textwidth}
\begin{figure}[H]
    \centering
    % \subfloat[\centering Memorization score]{{\includegraphics[width=0.49\textwidth]{figs/memorize_score_cor.png} }}%
    % \subfloat[\centering $k$-extractible]{{\includegraphics[width=0.49\textwidth]{figs/memorize_extra_cor.png} }}%
    \subfloat[\centering Memorization score]{{\includegraphics[height=1.45in]{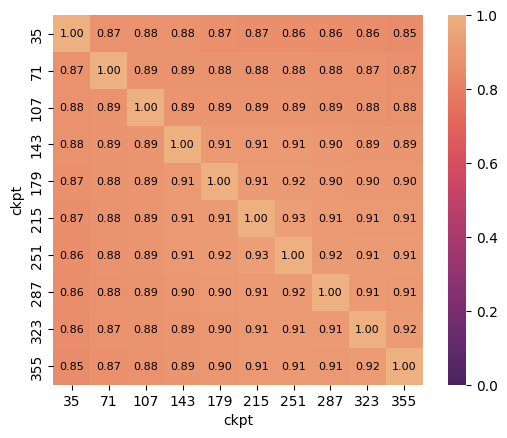} }}%
    \subfloat[\centering $k$-extractible]{{\includegraphics[height=1.45in]{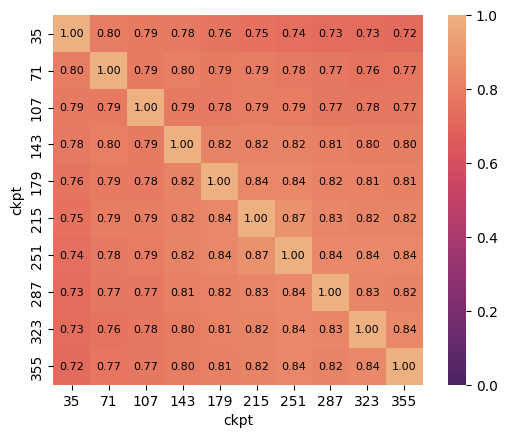} }}%
    \caption{The correlation of sequences in terms of memorization score and $k$-extractible between each checkpoints}%
    \label{fig:mem-corr}%
\end{figure}
% \caption{A figure with two subfigures}
% \label{fig:test}
%   \centering
%   \includegraphics[width=\linewidth]{figs/memorize_score_cor.png}
%   \includegraphics[width=\linewidth]{figs/memorize_extra_cor.png}
%   \captionof{figure}{placeholder}
%   \label{fig:test2}
\end{minipage}

\subsection{\analysis{}}\label{sec:analysis360}
Prior work such as Pythia~\cite{biderman2023pythia} has shown that an insightful study can be done by analyzing the intermediate checkpoints of a model. We hope \llm{} can also provide the community useful resources for both reference and research purposes.  To this end, we  release the initial version of the \href{https://github.com/LLM360/Analysis360}{\analysis{}} project, an organized repositories that analyze the model behavior on various aspects, including model characteristics and downstream evaluation results.

As an example of the analysis that can be performed over the set of model checkpoints, we conduct an initial study on memorization in LLMs.
Recent work \cite{carlini2021extracting, carlini2022quantifying} shows that LLMs may memorize a significant part of their training data, which can be extracted with appropriate prompting.
Such memorization not only raises privacy concerns in leaking private training data, but also downgrades the performance of LLMs if the training data contains unintended duplicates or peculiarities.
As we release all checkpoints and data, we can conduct a comprehensive analysis of memorization across the whole stage of training.

We adopt the \textit{memorization score} introduced in \cite{biderman2023emergent}, indicating the accuracy of tokens in the continuation of length $l$ with a prompt of length $k$,
\begin{align*}
    \mathrm{score}(k, l) = \frac 1l \sum_i^l \mathbf 1[S_{k+i} = G_{k+i}],
\end{align*}
where $S_{0:k+l}$ is the sequence from training data, while $G_{k:k+l}$ is the generated sequence with prompt $S_{0:k}$.
A \textit{memorized} or $k$-extractible \cite{carlini2021extracting} sequence has a memorization score of $1$.
Following \cite{biderman2023emergent, biderman2023pythia}, we conduct our experiments with $k = l = 32$.
We sampled $1000$ sequence from each of the $360$ data chunks, and use the first $64$ tokens of each sequence to conduct the following experiments.

We show the distribution of memorization scores for $10$ selected checkpoints in Figure \ref{fig:mem-overall}, and additionally annotate the percentage of $\mathrm{score}=1$. For every checkpoint, we only include the data chunks it has already been trained on.
From the result, we learn that
1) More than $1\%$ of the sequences are $32$-extractible from \amber;
2) \amber{} can memorize more sequences with the training going;
3) The spike at $\mathrm{score}=1$ indicates that \amber{} can memorize a much larger number of tokens than our preset threshold $32$ (consistent with prior work \cite{carlini2022quantifying, biderman2023emergent}).

We group the data chunks according to the selected checkpoints, and plot the memorization score on each data chunk group for each checkpoint in Figure \ref{fig:mem-data}.
We find that
1) \amber{} checkpoints memorize the latest seen data much more than previous data;
2) For each data chunk, the memorization score drops a bit with additional training, but keeps increasing afterwards.

We show the correlation between sequences in terms of memorization score or $k$-extractible in Figure~\ref{fig:mem-corr}.
We witness a strong correlation between the checkpoints.
% \gy{\cite{biderman2023emergent} reports many results on correlation. However, I think such correlation is mainly due to the data distribution itself, i.e., some of the data align well with the overall dataset, while others are not, causing a high correlation. It does not have much causal relationship from `memorization` of LLM. I think we may discard this result from the paper.}

\section{Summary and Take-home Messages}
In this section, we summarize the observations and a few take-home messages from our pre-training of \amber{} and \crystal, our initial modeling efforts in the LLM360 series. We understand that pre-training is a computationally daunting task that many academic labs or small organizations cannot afford to conduct. We hope that LLM360 can provide comprehensive knowledge, allowing users to understand what happens during LLM pre-training (\eg loss curve behaviors, how the evaluation metrics emerge, etc.) without the need to do so themselves. We also provide some potential use cases showing how researchers and developers can use LLM360 for their own projects.

\paragraph{Take-home Messages} Below we list a few of the lessons learned during our initial model training.

\begin{itemize}
    %\item The major progress in MMLU evaluation appears to occur mainly during the latter part of LLM pre-training. Observing our MMLU curves, the score initially decreases in the first half of model training before starting to increase. However, we are not yet certain whether this observation is specific to the 7$B$ model or a general phenomenon in LLM pre-training. We will continue to investigate this effect in our larger LLM pre-training efforts.
    \item In the pre-training of \amber{}, NaN losses were periodically observed, which may have been caused by certain random states, the training precision, or data quality issues. Some solutions include switching to a different random seed or skipping those data chunks. We notice some ``misbehaved'' data chunks can cause NaN loss regardless of when they are trained. In a preliminary experiment, we move the ``misbehaved'' data chunks to the end of the training but still observe NaN losses. %We will also provide these Recent research such as~\cite{zeng2023glm130b} suggest a few solution to this.
    \item %FSDP may not be the optimal option for the training system. 
    In the pre-training of \crystal{} and our subsequent LLM pre-training efforts, we observed that a hybrid and carefully tuned parallelism strategy—combining data, tensor-model, and pipeline (also referred to as 3D) parallelism strategies~\cite{narayanan2021efficient}—achieves better system throughput than FSDP, especially in distributed clusters with limited intra-node bandwidth.
    \item Data cleaning (and/or data quality filtering), along with data mixing ratios, are crucial aspects of LLM pre-training, as is the scheduling for various pre-training data categories (\eg CommonCrawl, Books, StarCoder, etc.). In \amber{} pre-training, we attempted to adhere as closely as possible to the hyperparameters used in LLaMA; however, our performance still lags significantly behind LLaMA's. A key omission in LLaMA's technical report is a detailed description of their exact pre-training dataset. Our carefully crafted \crystal{} pre-training dataset, which mixes English and coding data, achieves competitive performance with LLaMA on both the Open LLM Leaderboard and Code Evaluation benchmarks. We, along with the entire LLM open-source community, are diligently exploring the best approaches for data cleaning, data quality filtering, and determining optimal data mixing ratios, a pioneering effort exemplified by the DoReMi method~\cite{xie2023doremi}.
    %Data cleaning (and/or data quality filtering), the data mixing ratios are crucial aspects of LLM pre-training, and the pre-training schedule for various pre-training data categories (\eg CommonCrawl, Books, StarCoder, etc.). We attempted to adhere as closely as possible to the hyperparameters used in LLaMA; however, our performance still significantly lags behind LLaMA. A key omission in LLaMA's technical report is the detailed description of their exact pre-training dataset. Our carefully crafted pre-training \crystal{} dataset mixes English and coding data and achieves competitive performance on both Open LLM Leaderboard and Code Evaluation benchmarks. We, along with the entire LLM open-source community, are diligently exploring the best approaches for data cleaning, data quality filtering, and determining optimal data mixing ratios (a pioneering work along this direction is the DoReMi method~\cite{xie2023doremi}). 
\end{itemize}

\paragraph{Potential Use Cases of LLM360}
We describe a few potential use cases of LLM360 below.
\begin{itemize}
    \item One can conduct experimental studies at any stage of model training. As previously mentioned, the optimal data mixing ratio remains a significant open problem in LLM pre-training. However, it is nearly impossible to verify a specific mixing ratio by conducting full LLM pre-training. A more feasible approach is to adjust the data mixing ratios on the fly, \ie starting from an intermediate checkpoint, and either increasing or decreasing a specific data ratio from a particular category, \eg increasing the data weight in Wikipedia.
    \item For building domain-specific LLMs (\eg medical, finance, law, etc.), one may not necessarily want to start from the last pre-trained LLM checkpoint (which would make it more akin to fine-tuning). Instead, one can always pick one of the LLM360 checkpoints (\eg from 50\% of the pre-training stage) and resume the pre-training to obtain a domain-specific LLM.
    \item A lot of algorithmic approximation frameworks for efficient training require partially trained model weights~\cite{wang2021pufferfish,wang2023cuttlefish}. LLM 360 provides perfect model initializations for those methods.
\end{itemize}

\paragraph{\llm{} and Responsible Usage}\label{sec:responsible}
Given the wide-ranging applicability and high performance of LLMs, applications powered by them have the potential to deeply influence various aspects of life. Consequently, it becomes essential for all parties involved in the chain of production of LLMs to carefully manage the potential impact and risks associated with them. All stakeholders need to be informed of these implications and take necessary actions accordingly. 

We believe the transparent nature of the \llm{} initiative can help make the potential risks known to stakeholders. As one example, many risks associated with LLMs are related to certain forms of biases~\cite{risk_taxonomy}, such as the risk of social stereotypes, discrimination and exclusion, and the risk of under-representing certain languages or domains. By inspecting the exact training data and bias analysis (e.g. BOLD~\cite{Dhamala_2021}) in \analysis{}, stakeholders can have a thorough review of these risks before deploying the models. \llm{} can also help with risk mitigation. The project shares reproducible traces and exact data during LLM training, providing a reusable environment for researchers to conduct experiments to design better guardrails to contain potential risks.

We understand the importance of controlling the risk of LLMs and we are committed to further developing the \llm{} framework to foster responsible usage of LLMs. We would like invite the community to work with us, by sharing research results or by simply providing feedback.

%\williex{Hector: can work on this section? Add a section here on Safety and Responsibility! Talk about how our framework with improved transparency allows for better safety, and how it is ok to open source our models because they are worse than existing models (where only the final checkpoint is released)}
\section{Conclusion and Future Work}
\label{sec:conclusion}

In this paper, we introduce \llm{}, an initiative for comprehensive and fully open-sourced LLMs. Along with the first release of \llm{}, we released two 7B LLMs: \amber{} (an English general-purpose LLM) and \crystal{} (an LLM pre-trained specifically for code generation). In terms of artifacts, we released pre-training code, configurations, hyperparameters, intermediate model checkpoints, optimizer states, as well as the data sequence and data processing code. Our vision is to significantly advance and promote transparency within the open-source LLM pre-training community.

For future work, we are conducting a more detailed analysis on \amber{} and \crystal{}'s base models as well as their fine-tuned models. Detailed results will be released and discussed in their respective technical reports. Our team is also pre-training a much larger LLM, which will be fully released as soon as the pre-training is complete. Additionally, we will explore the optimal ratios for mixing different subsets in the pre-training datasets.
\subsection*{Acknowledgements}
We would like to thank Natalia Vassilieva, Joel Hestness, William Marshall, and Bhargav Kanakiya for their contribution to \crystal{} and support on the \llm{} project. We would also like to thank the MBZUAI and Cerebras team for providing and managing the computing infrastructure. 
%%%%%%%%%%%%%%%%%%%%%%%%%%%%%%%%%%%%%%%%%%%%%%%%%%%%%%%%%%%%

\bibliography{llm360}
\bibliographystyle{unsrt}
\end{document}